\pdfoutput=1

\documentclass[11pt]{article}

\usepackage[final]{acl}
\usepackage{times}
\usepackage{latexsym}
\usepackage[T1]{fontenc}
\usepackage[utf8]{inputenc}
\usepackage{microtype}
\usepackage{inconsolata}
\usepackage{graphicx}
\usepackage{amsmath}
\usepackage{amssymb}

\usepackage{listings}
\usepackage{threeparttable}
\usepackage{subcaption}
\usepackage{paralist}
\usepackage{float}
\usepackage{comment}

\usepackage{algorithm}
\usepackage{algorithmic}
\usepackage{subcaption}
\usepackage{multicol}

\definecolor{darkred}{RGB}{139,0,0}
\definecolor{lightred}{RGB}{255,99,71}
\definecolor{amber}{RGB}{255,191,0}
\definecolor{lightgreen}{RGB}{144,238,144}
\definecolor{darkgreen}{RGB}{34,139,34}

\definecolor{navyblue}{RGB}{0,0,128}

\newcommand{\ie}{i.e.,\ }
\usepackage{booktabs}
\usepackage{multirow}
\usepackage{array}
\usepackage{rotating}
\usepackage{fp}
\usepackage{makecell} 
\usepackage{tabularx}
\usepackage{listings}
\usepackage{soul}  
\usepackage[textsize=small]{todonotes}

\usepackage[inline]{enumitem}

\usepackage{url}

\usepackage[most]{tcolorbox} 

\usepackage{listings} 

\newcommand{\inlineicon}[2][0em]{%
    \raisebox{#1}{\includegraphics[height=0.8em]{#2}}%
}

\title{Generating Place-Based Compromises Between Two Points of View}
\author{
  Sumanta Bhattacharyya\thanks{Work done during internship at Toyota Research Institute.} \\
  University of Illinois Chicago \\
  \texttt{sbhatt54@uic.edu}
  \And
  Francine Chen \\
  Toyota Research Institute \\
  \texttt{francine.chen@alum.mit.edu}
  \And
  Scott Carter \\
  Toyota Research Institute \\
  \texttt{scott.carter@tri.global}
  \AND
  Yan-Ying Chen \\
  Toyota Research Institute \\
  \texttt{yan-ying.chen@tri.global}
  \And
  Tatiana Lau \\
  Toyota Research Institute \\
  \texttt{tatiana.lau@tri.global}
  \And
  Nayeli Suseth Bravo \\
  Toyota Research Institute \\
  \texttt{nayeli.bravo@tri.global}
  \AND
  Monica P. Van \\
  Toyota Research Institute \\
  \texttt{monica.van@tri.global}
  \And
  Kate Sieck \\
  Toyota Research Institute \\
  \texttt{kate.sieck@tri.global}
  \And
  Charlene C. Wu \\
  Toyota Research Institute \\
  \texttt{charlene.wu@tri.global}
} 

\renewcommand{\todo}[1]{{\color{red} TODO: {#1}}}

\begin{document}
\maketitle

\begin{abstract}


Large Language Models (LLMs) excel academically but struggle with social intelligence tasks, such as creating good compromises. 
In this paper, we present methods for generating empathically neutral compromises between two opposing viewpoints.
We first compared four different prompt engineering methods using  Claude 3 Opus and a dataset of 2,400 contrasting views on shared places. 
A subset of the generated compromises was evaluated for acceptability in a 50-participant study. We found that the best method for generating compromises between two views used external empathic similarity between a compromise and each viewpoint as iterative feedback, outperforming standard Chain of Thought (CoT) reasoning. The results indicate that the use of empathic neutrality improves the acceptability of compromises.
The dataset of generated compromises was then used to train two smaller foundation models via margin-based alignment of human preferences, 
improving efficiency and removing the need for empathy estimation during inference. 

\end{abstract}
\section{Introduction}
Living in liberal, plural societies requires compromise and consensus. While these two components are essential for healthy deliberation and decision-making, they are not the same. For example, while multiple parties in one household could prefer different types of meat for dinner (e.g., chicken, beef, lamb), they might come to a consensus that they all want meat for dinner, but they may also compromise by agreeing to delay eating beef and lamb for another time and eating chicken tonight, because there is only chicken in the refrigerator. 

Recent research shows that LLMs can promote democratic deliberation by fostering consensus (developing statements different participants can agree on) and reducing polarization, often outperforming human mediators in clarity and inclusiveness \cite{tessler2024ai}. Additionally, \citet{bakker2022fine} show that fine-tuning LLMs to generate consensus statements from diverse individual preferences enhances group acceptance of the statements more efficiently than traditional negotiation processes. 

However, these examples concern using LLMs to develop \textit{consensus} statements across broad viewpoints. Developing \textit{compromises} between different and opposing viewpoints, on the other hand, requires a greater degree of social intelligence that LLMs may lack \cite{xu2024academically}.
Although LLMs can produce empathetic responses \cite{welivita2021large,sabour2022cem,majumder2020mime} and can generate consensus statements, they have not been used to generate neutral compromises. 
Generating compromises are fundamentally different, as successful compromises (i.e., acceptable to both parties) require proposing resolutions that make both parties better off compared to in the absence of any compromise \cite{van2012makes}. This thus requires attending to a solution space that allows for differing viewpoints to both find positives compared to their original stances.
Recent studies show that while LLMs can analyze disputes and generate intervention messages, their effectiveness as neutral mediators remain inconsistent \cite{tan2024robots}. Similar concerns arise in humanitarian frontline negotiations, where neutrality and sensitivity are critical. Interviews with practitioners highlighted that, although LLMs can synthesize large amounts of information, they may be biased, which can lead to mixed outcomes when supporting neutral negotiation processes \cite{ma2024using}.


\begin{figure*}[t]    
    \centering
    \includegraphics[width =\linewidth]{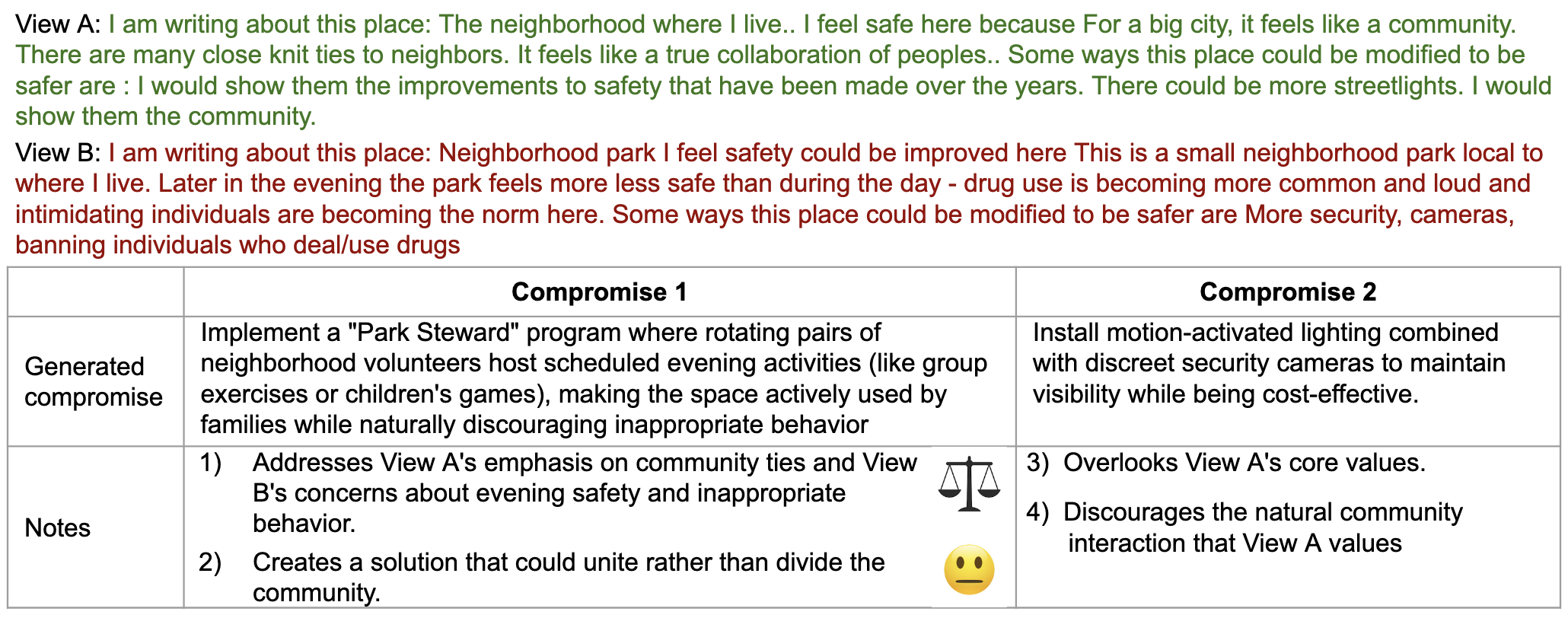}
    \caption{Example showing two contrasting views, View\_A and View\_B, two compromises  generated for the two views using prompt engineering, and notes about each compromise. Compromise 1 balances both views and is empathicly neutral, while Compromise 2 does not.}
    \label{fig:datapoint}
\end{figure*}

Inspired by these earlier works, 
this paper explores whether LLMs can generate  
neutral compromises for contrasting views (i.e., generate compromises that do not favor either view). This ability would be valuable in scenarios where impartiality 
is crucial, such as collaborative decision-making and conflict resolution. 
Particularly, we investigate the following research questions:
\\\textbf{RQ1:} Can prompt engineering be used to generate compromises that are empathically neutral between two diverse viewpoints (about a place)?
\\\textbf{RQ2:} Can smaller, open-source LLMs replicate empathically neutral compromise generation capabilities based on a larger, proprietary model, and remove the need for explicit empathy estimation during inference?%

To address these questions, we propose a task that focuses on generating compromises from a pair of contrasting views (Figure \ref{fig:datapoint}). As a first exploration of this task, we use pairs of contrasting positive and negative viewpoints about safety and welcomingness of public spaces from \citet{chen2025empathy}. The positive views (view\_A) are about feeling either safe or welcome  and the negative views (view\_B) are about feeling less safe or excluded. Each viewpoint includes a participant's proposal for improving the safety or welcomingness of a place.
In our task, generated compromises for improving safety or welcomingness  between a pair of positive and negative views should: \begin{inparaenum}[(i)] \item consider both suggestions and \item  be empathically neutral\end{inparaenum}. 
Since greater empathy helps people to understand each other's opinions and viewpoints, empathy is important to compromise, which requires concessions from the parties involved. While individual parties agree to a compromise because they would be worse off without it  \cite{van2012makes}, if two views are judged to 
be similarly empathetic towards a compromise, or \textit{empathically neutral},
the probability of finding a suggested compromise acceptable should naturally increase. 


To address RQ1, we explore several
prompt-based controls to guide the generation process. 
We investigated guiding the model toward empathic neutrality by providing explicit feedback on the empathic similarity \cite{shen2023modeling} between each of the two views and the generated compromise (Section \ref{subsec:sim_model}). When the empathic similarity scores across both views are balanced, the generated compromise is considered \textit{empathically neutral}. 
A 50-participant human evaluation of a subset of the compromises was conducted to assess the acceptability of the different compromise generation approaches and showed that compromises generated by the model informed with empathic neutrality was highly preferred.

To address RQ2, 
we use the compromises generated in RQ1 to explore an alignment technique to teach empathetically neutral compromise generation abilities to smaller, pretrained language models~\cite{liu2023aligning,huang2024antidote} Motivated by how humans internalize social norms not just by observing conforming behavior, but by distinguishing instances of norm conformation from instances of norm violation ~\cite{bicchieri2005grammar} and fair and unfair outcomes~\cite{fehr2004social}, we use NCE-based alignment to explicitly contrast validated compromises against inadequate alternatives.
Our experiments show that the approach improved the quality of smaller LLM-generated compromises over fine-tuning alone, which only exposes models to acceptable compromises.  

In this paper we present a method that uses empathic neutrality to enhance prompt engineering to generate compromises. These compromises are then used, together with poorer quality compromises, to teach an alignment model to align smaller models. 
In the following sections, we describe the different prompt-based models that we investigated and the automatic and human evaluation of the generated compromises. We also describe the method for aligning smaller language models to generate empathically neutral compromises that we investigated.

\label{sec:int}
\section{Related work}

 LLMs excel in academic tasks (GPT-4 achieving 92.1\% on MATH) but exhibit limitations in social intelligence, scoring only 54.4\% on SESI benchmark \cite{xu2024academically}. Similar limitations appear in interactive gaming contexts, where LLMs perform approximately 20\% below human baseline in theory of mind tasks \cite{liu2024interintent}. Studies from First-person perspective confirm that despite possessing basic theory of mind capabilities, LLMs show considerable limitations in managing complex social interactions compared to human performance \cite{hou2024entering}. Current evaluation methods include traditional psychological assessments like ToMi \cite{le2019revisiting} and specialized datasets such as SocialIQA \cite{sap2019socialiqa}, SocKET \cite{choi2023llms}, and SECEU \cite{wang2023emotional}. However, newer benchmarks like SOTOPIA \cite{zhousotopia} and EmoBench \cite{sabour2024emobench} face limitations in reflecting real-world interactions. Prior work generated consensus statements using supervised fine-tuning and re-ranking \cite{bakker2022fine}, or involved humans in selecting common ground \cite{tessler2024ai}. We instead generate multiple compromise suggestions from open-ended prompts, aiming for balanced empathy across differing views without human selection or re-ranking.

\textbf{Synthetic datasets} offer a promising solution for enhancing LLMs' social capabilities \cite{ghanadian2024socially, hassan2024automated, balog2024towards, gabriel2024development}. These datasets leverage LLMs' generative abilities \cite{dankar2021fake} while ensuring diverse representation \cite{yamagishi2024utrad}. Advanced prompt engineering techniques, particularly using GPT-4 or Claude \cite{achiam2023gpt, anthropic2023introducing}, have improved dataset quality \cite{hikov2024information,shi2023safer}. Chain of Thought (CoT) approaches \cite{wei2022chain, feng2024towards, chu2023survey} show enhanced reasoning capabilities \cite{shao2023synthetic}, with significant improvements in specific tasks \cite{nong2024chain}. Structured Chain-of-Thought (SCoT) \cite{sultan2024structured} further improves accuracy in document-grounded QA conversations. Recent development suggest using parameterized soft prompts instead of traditional hard-prompting \cite{desalvo2024no}.

\textbf{Model alignment} with human preferences has progressed from Reinforcement Learning from Human Feedback (RLHF) \cite{kaufmann2023survey,ouyang2022training} to more efficient alternatives \cite{liu2023aligning, kim2024aligning, wang2024comprehensive}. While RLHF aligns models with complex human values through reward models and policy optimization \cite{song2024preference, sun2023aligning, yuan2024rrhf}, it remains resource-intensive and complex \cite{schulman2017proximal, wang2024arithmetic, hong2023cyclealign}. Direct Preference Optimization (DPO) \cite{rafailov2024direct} simplifies training by removing the reward model and optimizing the likelihood of preferred responses. Extensions such as reward model distillation \cite{fisch2024robust} and KL regularization \cite{kullback1951information, wesego2024scorebased, azar2024general} further enhance stability. DPOP \cite{pal2024smaug} adds a penalty to maintain preference alignment with the reference model. InfoNCA and NCA \cite{gutmann2010noise, chen2024noise} adapt DPO for multi-response settings by optimizing absolute likelihoods. Finally, methods like TODO \cite{guo2024todo} and RPO \cite{yin2024relative} introduce ternary ranking and contrastive weighting to capture finer-grained preference signals.

This study focuses on aligning model outputs in the context of neutral compromise generation. Prompt engineering is guided by the empathic similarity between viewpoints and generated compromise, and this setup is used to generate synthetic training data. This data is used for fine-tuning and aligning pretrained LLM to generate these compromises.

\label{sec:bgnd}
\section{Compromise Generation using Empathy-Informed LLM}
\label{sec:cd}

This section describes the dataset used for our experiments
and the generation of candidate compromise sentences through diverse prompt engineering techniques.

\subsection{ContrastingViews Dataset}
\label{subsec:dc}

\begin{figure*}[!ht]
\centering
\includegraphics[width=2\columnwidth]{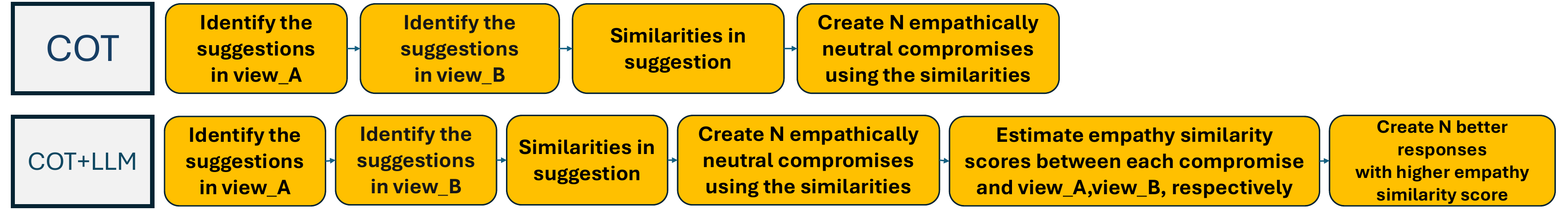}
\caption{ Prompt engineering strategies for collected
compromise generation (a) basic CoT approach (b) CoT+LLM}
\label{fig:allprompts}
\end{figure*}

\begin{figure*}[t]    
    \centering

\includegraphics[width=\textwidth,keepaspectratio]{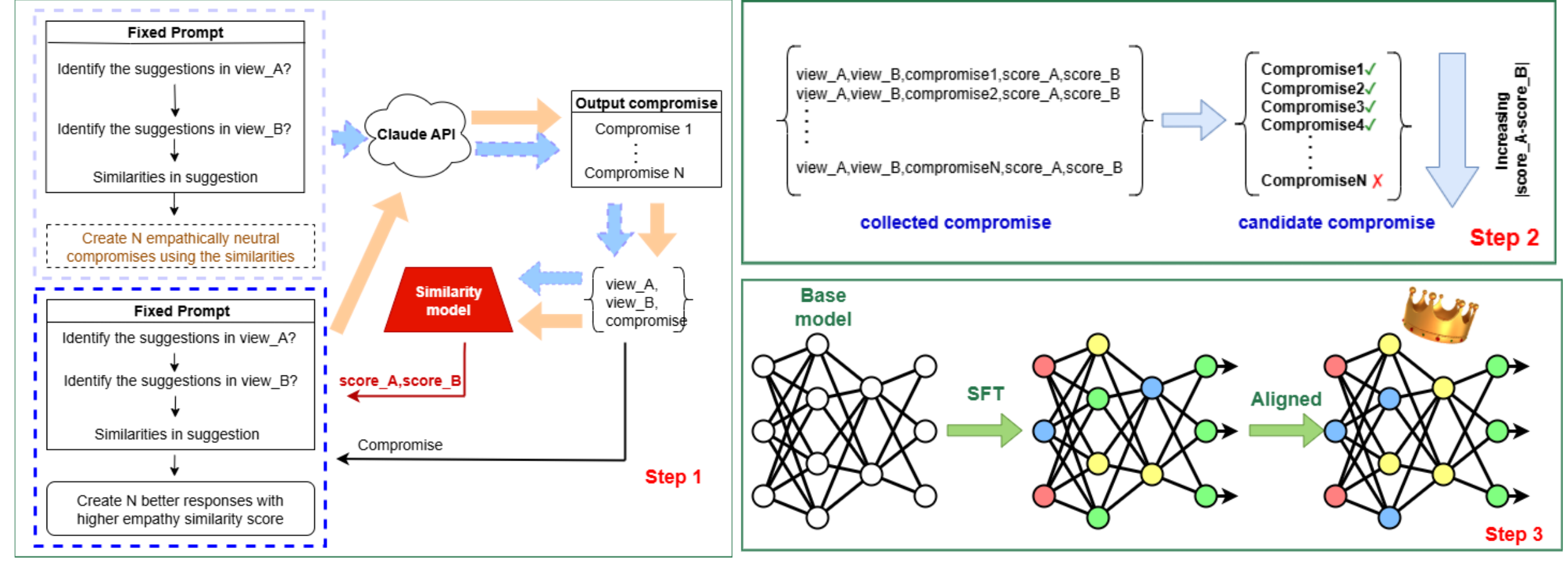}    
    \caption{Overview of the full pipeline used in our proposed method. \textbf{Step 1} depicts the iterative prompt engineering procedure used by CoT+Feedback to generate collected compromises guided by a similarity model (see Section \ref{subsec:gcc}). Two types of arrows illustrate the refinement iterations: \inlineicon[-0.2em]{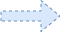} indicates the initial prompt flow used to generate collected compromises, while \inlineicon[-0.2em]{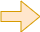} represents subsequent iterations that further refine the collected compromises.  \textbf{Step 2} filters  candidate compromises based on neutrality (see Table \ref{tab:response-distribution}). \inlineicon[-0.2em]{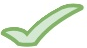} denotes an accepted candidate compromise and \inlineicon[-0.2em]{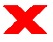} denotes less neutral, hence not accepted. \textbf{Step 3} presents the alignment process (see Section \ref{sec:meth}). }
    \label{fig:flow}
\end{figure*}

We adapt the public EmpathyFromPerspectives dataset \citet{chen2025empathy} composed of pairs of contrasting viewpoints by human participants about feeling safe/less safe and welcome/excluded in public spaces (see Appendix \ref{sec:dcm} for a summary). Each pair of viewpoints includes: view\_A, which is either a safe or welcome viewpoint and a corresponding view\_B, which is either a less safe or excluded viewpoint. Each viewpoint is composed of a place description, reasons for the viewpoint, and suggested improvements. We randomly selected 1200 safe/less safe and 1200 welcome/excluded pairs for our generation experiments, and refer to this subset as the \textit{ContrastingViews} dataset.

\subsection{Empathic similarity model}
\label{subsec:sim_model}

We adopt the empathic similarity framework from~\citet{shen2023modeling}, which computes an empathic similarity score $sim(f_{\theta}(s_1), f_{\theta}(s_2))$ between pairs of stories $(s_1, s_2)$, where higher scores indicate greater alignment in empathic content. While~\citet{shen2023modeling} used Sentence-BERT~\cite{reimers2019sentence}, we replace it with the e5-large model~\cite{wang2022text} for improved sentence representation.
To adapt the similarity model to our scenario, we combined 2,000 annotated story pairs from Shen's  EMPATHICSTORIES dataset \cite{shen2023modeling} 
with 1,000 story pairs from the ContrastingViews dataset. 
Prior work~\cite{schopf2023efficient,hupkes2022state} has shown that such pretrained encoders can support semantic generalization even when exposed to mismatched discourse styles, such as narrative and structured text.
The combined dataset of 3,000 pairs is used to train the similarity model with a 75/5/20 train/validation/test split.
During inference, this model estimates empathic similarity between a generated compromise and each of its corresponding pair of viewpoints. 

\subsection{LLM generation of compromise data }
\label{subsec:gcc}

We aim to generate compromises for each given pair of positive and negative views for a place. Each view consists of  \begin{inparaenum}[(i)] \item description of the place \item reason of the view \item suggestions of improvement\end{inparaenum}. In order to use all this information as a context for feeding into an LLM, we follow a prompt template similar to the Alpaca prompt \cite{alpaca}. Recent work suggests that proprietary LLMs excel at synthetic dataset generation \cite{cascante2023going,lei2023s3eval,shi2023safer} with the help of task specific prompt based strategies, providing an alternative to human data collection. We use Claude 3 Opus \cite{anthropic2023introducing} for generating compromise data. 

Using a single compromise as the target constrains model generalization capabilities \cite{gong2019diversity}. We generated multiple candidate compromises per view pair, enhancing the model's ability to learn diverse resolution patterns. Additionally, we observed that single prompts were inadequate (see section~\ref{sec:sprompt} of the appendix) for our task as they frequently focus on only one perspective, which necessitated our exploration of more sophisticated prompt engineering techniques. All three CoT-based strategies share a common structural principle designed to promote balance between the two viewpoints. Each begins by separately extracting the suggestions from view\_A and view\_B before identifying commonalities. This explicit decomposition ensures that both perspectives are represented in the model's reasoning context prior to compromise generation, preventing the model from anchoring on one viewpoint, a failure mode we observed frequently with single prompts (Section~\ref{sec:failure}). The strategies differ in how they reinforce this balance using empathic similarityafter initial generation, as described below.

\textit{Chain of Thought (CoT)}: We initially used CoT \cite{chu2024navigate, wei2022chain} to identify suggestion similarities and derive compromises, as illustrated in Figure \ref{fig:allprompts}. This approach, however, proved ineffective, the generated text either emphasizing a single suggestion too heavily or extending beyond their intended scope, ultimately leading to less suitable compromises. 

\textit{CoT+LLM}: 
LLMs have demonstrated the potential to refine their outputs through self-generated feedback, closely aligning with human refinement techniques \cite{madaan2024self}. In this approach, we integrated LLM-based self-evaluation scores (Figure \ref{fig:allprompts}). These scores measure empathic similarity between the generated compromise and each view, ideally achieving high similarity for both perspectives with minimal discrepancy. Although self-evaluation scores showed improvement, we observed negligible gains in actual compromise quality, suggesting that these scores may not reliably enhance response quality.


\textit{CoT+Feedback}: To provide more reliable feedback for refining responses than LLM-based self-evaluation, we utilize empathic similarity scores from the trained empathic similarity model (Section \ref{subsec:sim_model}). 
Note that while view-view pairs are used to train empathic similarity, for CoT+Feedback, the empathic similarity model is used to score a novel generated compromise against each view.
The 480 pairs of views used for evaluation of all prompting models were separate from the 1,000 ContrastingViews pairs used to train the similarity model. 
The predicted scores, along with generated responses (step 1 in Figure \ref{fig:flow}), serve as iterative feedback to improve response quality. Across iterations, both response quality and empathy similarity scores show consistent improvement, indicating greater utility for refinement than self-evaluation alone.\\
Please refer to Appendix~\ref{sec:aprompt} for the qualitative analysis of the generated collected compromises across different prompting strategies.

\subsection{Candidate compromise selection}
For each prompting strategy, four compromises are collected per view pair (resulting in $4 \times 4 = 16$ compromises per view pair across four prompting strategies: Single Prompt, CoT, CoT+LLM and CoT+Feedback). From this set of 16, the top four responses are selected as candidate compromises (step 2 in Figure \ref{fig:flow}) by ranking them according to an estimate of empathic neutrality (see Table \ref{tab:response-distribution}). Neutrality is evaluated using the similarity model to compute the empathic similarity between the generated compromise and both view\_A and view\_B, denoted as \textbf{score\_A} and \textbf{score\_B}, respectively. Then empathic neutrality is computed as the absolute difference in empathic similarity $|score_A - score_B|$  In an ideal state of empathic neutrality, the absolute difference between these scores approaches zero.

\begin{table}[!htbp]
\centering
\small
\setlength{\tabcolsep}{4pt} 

\begin{tabular}{|l|c|c|c|c|}
\hline
\textbf{Topic} & SP & CoT & CoT+LLM & CoT+FB \\
\hline
Welcome & 0.75\% & 24.70\% & 28.45\% & 46.10\% \\
\hline
Safe & 0.83\% & 22.09\% & 26.65\% & 50.43\% \\
\hline
\end{tabular}

\caption{Distribution of candidate compromises across prompting strategies for each topic.}
\label{tab:response-distribution}
\end{table}

Table \ref{tab:response-distribution} shows the distribution of candidate compromises across all strategies. The CoT+Feedback approach contributed the highest number of desired candidate compromises, while the single prompt yielded the lowest, demonstrating the effectiveness of our method.

\section{Human Evaluation of Compromise Methods}
\label{sec:humaneva}

The previous evaluation indicates how well the generated compromises achieve empirical empathic neutrality as measured by comparing the empathic similarity of two opposing viewpoints. We next conduct a human evaluation.
\subsection{Evaluation Methodology }

For this, we asked 50 humans to rate compromises generated by the Single Prompt, \textit{COT}, and \textit{COT+Feedback} methods. The participants were shown pairs of statements about modifications proposed for a place, where one statement was written by someone who thought positively about the place and the other statement was written by someone who felt negatively about the place. Instead of asking directly how fair a compromise was, which could be biased by the rater's personal perspective and from cognitive priming and carryover effects of viewing two viewpoints sequentially \cite{epley2006anchoring}, a rater was asked to take the viewpoint of either the person who felt positively or who felt negatively for all pairs of statements presented (assigned viewpoint). The number of raters taking the positive and negative viewpoints was balanced. 

For each pair of statements, raters were asked to rate on a scale of 1-100 the acceptability of each of five compromises from their assigned viewpoint. One of the ``compromises'' was the positive or negative statement that they were not to identify with (`Opposing View'). The four other compromises were generated: one by Single Prompt, one by \textit{COT}, and two by the \textit{COT+Feedback} method. Each participant rated the compromises for five pairs of statements, providing ratings from the positive and negative viewpoints for a total of 100 pairs of statements and 400 generated compromises. 
For details about the collection of compromise ratings, see Appendix \ref{app:humanEval}.

\subsection{Preference Results }
\label{sec:prefResults}

Table~\ref{tab:humanstudy} shows the percentage of time that each of the compromise methods was the top choice (First Preference) and second choice (Second Preference). We observed that COT+Feedback compromises were most preferred (COT+Feedback 1: 37\% first, 33\% second; COT+Feedback 2: 40\% first, 29\% second), outperforming CoT (18\%, 24\%) and Single Prompt (5\%, 13\%). The opposing view was rarely preferred (0\%, 1\%).

To verify the statistical robustness of these preferences, we performed three tests on participant First Preferences: (1) a bootstrap analysis (10,000 iterations) to compute 95\% confidence intervals (2) Wilcoxon signed-rank tests by user to compute the significance of the different methods against the Single Prompt method and (3) permutation tests by on per-item ranks to also compute the significance of the different methods against the Single Prompt method. The results shown in Table~\ref{tab:human_eval} indicates that the preference for COT+Feedback is statistically significant compared to the Single Prompt baseline ($p \le 0.002$) for both the Wilcoxon and Permutation tests. 
In contrast, the standard COT method (18\%) was not found to be statistically distinguishable from the baseline ($p > 0.15$)

\begin{table}[ht]
\centering

\setlength{\tabcolsep}{6pt} 
\resizebox{0.85\columnwidth}{!}{%
\begin{tabular}{lcc}
\toprule
Method & \begin{tabular}[c]{@{}c@{}}First\\ Preference (\%)\end{tabular} & \begin{tabular}[c]{@{}c@{}}Second\\Preference (\%)\end{tabular} \\
\midrule
Opposing View\ & 0 & 1 \\
Single Prompt & 5 & 13 \\
COT & 18 & 24 \\
COT+Feedback 1 & 37 & 33 \\
COT+Feedback 2 & 40 & 29 \\
\bottomrule
\end{tabular}%
}
\caption{Preference distribution for user study (\%).  
For Single Prompt and \textit{CoT}, one randomly selected compromise was used in evaluation. 
For CoT+Feedback 1 and 2, two randomly selected compromises from the \textit{CoT+Feedback} method were used. 
Opposing View represents the opposing viewpoint to the person doing the rating.}
\label{tab:humanstudy}
\end{table}

\begin{table}[t]
\centering
\small
\setlength{\tabcolsep}{2.5pt}
\renewcommand{\arraystretch}{1.05}
\begin{tabular}{@{} l c c c @{}}
\toprule
\textbf{Method} &
\makecell{\textbf{First Pref.\ \%}\\\textbf{(95\% CI)}} &
\makecell{\textbf{Wilcoxon}\\\textbf{$p$}} &
\makecell{\textbf{Perm.}\\\textbf{$p$}} \\
\midrule
Opposing View   & 0.0\,\% [0.0--1.5]    & 0.128 & 0.144 \\
Single Prompt   & 5.0\,\% [2.8--8.2]    & --    & --    \\
CoT             & 18.0\,\% [13.7--23.2] & 0.157 & 0.188 \\
\textbf{CoT+FB\,1} & \textbf{37.0\,\% [31.1--42.9]} & \textbf{0.002} & \textbf{0.002} \\
\textbf{CoT+FB\,2} & \textbf{40.0\,\% [34.1--46.2]} & \textbf{0.001} & \textbf{0.002} \\
\bottomrule
\end{tabular}
\caption{Preference distribution with statistical significance. 95\% CIs via bootstrap (10k). $p$-values compare each method vs.\ \textit{Single Prompt}.}
\label{tab:human_eval}
\end{table}



These results indicate that the COT+feedback method significantly outperforms both COT only and single-prompt approaches, 
supporting its use for empathically neutral compromise generation.






\section{Aligning Foundation Models for Empathically Neutral Compromise Generation }
\label{sec:meth}

The prompt engineered pipeline described in Section 3 demonstrates that external empathic feedback improves compromise quality. However, this pipeline relies on iterative similarity scoring at inference time, requiring access to both a proprietary LLM and a trained similarity model. In human mediation training, practitioners learn to generate fair proposals not by consulting external fairness metrics in real time, but by building internal representations of what balanced perspective-taking feels like. We pursue an analogous goal computationally: aligning smaller, pre-trained language models so that empathically neutral compromise generation becomes an intrinsic capability rather than an externally guided one.

Instead of using standard fine-tuning to teach a model to imitate target outputs, we use an approach that teaches a model to distinguish good compromises from poor ones. This choice is motivated by cognitive research on social norm acquisition showing that humans learn evaluative standards through contrastive experience, e.g., comparing norm-conforming behavior against violations~\cite{fehr2004social,bicchieri2005grammar}.

We adopt Noise Contrastive Estimation (NCE)~\cite{gutmann2012noise,chen2024noise}, which directly optimizes the absolute likelihood of preferred outputs without requiring a separate reference model. This is important for our task because the model must retain broad world knowledge to generate contextually appropriate compromises. In contrast, alignment methods that optimize likelihood ratios, such as DPO~\cite{rafailov2024direct}, risk suppressing the absolute likelihood of even preferred outputs, degrading generation quality.

\subsection {NCE based alignment}
\label{subsec:nces}

We train an NCE model using the candidate compromises preferred by human evaluators (Section \ref{sec:prefResults}) as positive examples 
to align the model with verified human social preferences. The negative examples are the fine-tuned base model's outputs, which serve as contrasting cases.



For training efficiency, all parameters are frozen except the last three layers. 
Let s\_target denote the sum of log-probabilities that the model assigns to the tokens of the candidate compromise, and let s\_hypo denote the corresponding sum for the fine-tuned base model's own generated output. The NCE loss (Equation~\ref{eq:nce}) consists of two terms:


\vspace{-2em}
\begin{equation}
\mathcal{L}_{\text{nce}} = -\log\left(\frac{1}{1+e^{-s_\text{target}}}\right) - \log\left(\frac{1}{1+e^{s_\text{hypo}}}\right)
\label{eq:nce}
\end{equation}
\vspace{-0.3em}
The first term encourages the model to increase s\_target, raising the likelihood of producing the validated compromise. The second term encourages the model to decrease s\_hypo, lowering the likelihood of generating its current, less neutral output. Together, the two terms train the model to discriminate between good and poor compromises, internalizing the preference signal that was previously provided externally by the similarity model.

\section {Experimental setup on alignment}
\label{sec:exp}

\textbf{Datasets:} We use the ContrastingViews dataset together with the synthetic compromises generated by CoT+Feedback, described in Section \ref{sec:cd}, to validate our alignment targets. Our smaller models are trained, when given a pair of opposing views, to mimic the compromises, whose quality was evaluated by humans in Section \ref{sec:humaneva}. Thus our alignment process optimizes towards a socially valid objective. The ContrastingViews dataset together with the genereated compromises were split into training (75\%), development (5\%), and test (20\%) subsets.

\textbf{Setup:} For all the experiments, we use Llama-3.1 8B ~\cite{dubey2024llama} and  mistral-7B instruct ~\cite{jiang2023mistral} as base models (referred as Llama and Mistral-7b). We initially fine-tune the base models for one epoch and use this fine tuned model for further alignment tasks. We implemented a linear warm-up and cosine decay scheduler to dynamically adjust the learning rate with an initial learning rate of $3e^{-5}$. Optimization is performed using the Adam optimizer with $\beta_1 = 0.90$ and $\beta_2 = 0.99$. Through empirical tuning, we found that a weight margin value of 10 was optimal for task-based loss alignment. We train for 8 and 12 epochs for the NCE-based alignment and task- loss based alignment, respectively.

\textbf{Evaluation metrics:}  
We employ a multi-faceted evaluation strategy: \begin{enumerate*}[label=(\roman*)]
\item \textbf{Generation Quality:} We report ROUGE scores~\cite{lin2004rouge} to measure semantic alignment with the human-validated candidate compromises (see Table~\ref{tab:rouge}).
\item \textbf{Empathic Neutrality:} We measure the "neutrality gap" ($|score_A - score_B|$), where lower values indicate better balance between viewpoints.
\end{enumerate*}

\section{Evaluation of alignment}
\label{sec:eva}

\begin{table*}[htbp]
\centering
\small
\begin{tabular}{|l|c|c|c|c|}
\hline
\multirow{2}{*}{Model  } & 
\multicolumn{2}{c|}{Llama} & \multicolumn{2}{c|}{Mistral-7b} \\
\cline{2-5}
& ROUGE-1($\uparrow$) & ROUGE-L($\uparrow$) & ROUGE-1($\uparrow$) & ROUGE-L($\uparrow$) \\
\hline
Base Model &0.164 &0.107 &0.160 &0.105 \\
\hline
Base Model + Sampling (best of 5) &0.193 &0.151 &0.191 &0.127 \\
\hline
Base Model + Finetuning (FT) &0.255 &0.186 &0.256 &0.185 \\
\hline
\textbf{Base Model + FT+ NCE} &0.315 &0.216 &0.318 &0.234 \\
\hline
\end{tabular}
\caption{Quantitative analysis on the test data. We calculated ROUGE score for the generated compromises across different baselines. Both of the aligned model outperforms other baselines. }
\label{tab:rouge}
\end{table*}
\subsection{Quantitative Evaluation}
As shown in Table \ref{tab:rouge}, initial generations from pre-trained baseline models perform poorly in producing high-quality compromises (\textbf{row 1}). 
To explore sampling-based improvement, we generate five samples using multinomial sampling and select the best one based on the evaluation metrics (\textbf{row 2}). However, sampling still fails to produce suitable compromises. Fine-tuning these models for a single epoch improves performance (\textbf{row 3}), but remains insufficient. Although DPO is a standard baseline, we excluded it from this generation quality analysis because our initial experiments revealed it caused significant catastrophic forgetting (detailed in Appendix~\ref{subsec:cata}). Given that social reasoning requires retaining broad world knowledge, we focused on our proposed alignment algorithms. This leads to consistent and substantial improvements in ROUGE scores across both Llama and Mistral-7b, thereby demonstrating the effectiveness of alignment-based training objectives for the compromise generation task (\textbf{rows 4 and 5}). 

In Appendix \ref{sec:task_loss}, we describe our investigation of a task-based alignment objective with contrasting examples, which also lead to improved ROUGE scores. And we  explored conditioning the generation on participant demographics (\ie age, gender) to test if explicit social context improves neutrality. However, our experiments yielded no statistically significant performance gains 
(Section~\ref{sec:demo}).

\subsection{Empathic Neutrality}

\begin{figure}[h]
\centering
\includegraphics[width=\columnwidth]{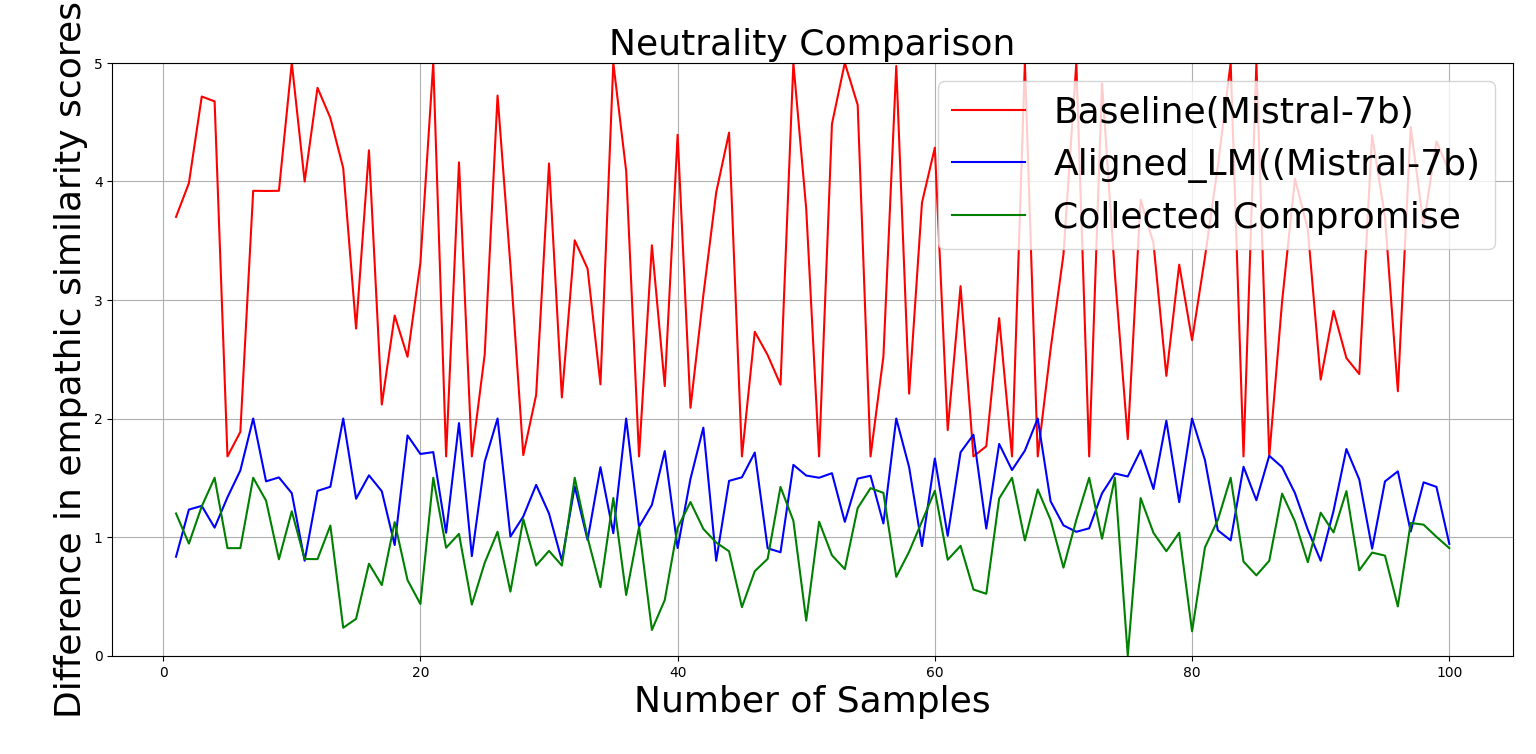}
\caption{ Difference between score\_A and score\_B: lower values indicate higher neutrality. The baseline model (Mistral-7b) produces less neutral compromises, resulting in a higher score difference. In contrast, the compromises generated via iterative prompt engineering achieve the lowest difference, reflecting the highest neutrality. After aligning the base model, it produces significantly more neutral compromises than the baseline.}
\label{fig:error_comparison}
\end{figure}

Improvements in ROUGE scores do not necessarily reflect increased empathic neutrality. As defined in Section 6, we quantify neutrality as the absolute difference between empathic similarity scores ($|score_A - score_B|$). Using 100 randomly sampled pairs of contrasting views from the test set, we analyze these differences across models. As shown in Figure \ref{fig:error_comparison}, the baseline (Mistral-7b) produces compromises with poor neutrality. In contrast, our method (aligned\_LM) using task-loss based alignment consistently generates compromises with smaller empathic similarity gaps, indicating improved neutrality. The candidate compromises 
serve as an upper bound for neutrality. Notably, aligned\_LM approaches this level of neutrality, bridging the gap between baseline generated compromise and prompt-generated compromise performance. 
Appendix~\ref{sec:quality}  provides a qualitative analysis of compromise generation across the baselines. \par



\section{Conclusion and Future Work}
Our work presents an initial framework for improving the social intelligence of LLMs for the novel task of empathically neutral compromise generation. We propose and conduct human evaluation of the effectiveness of several prompting strategies, and show that use of empathic similarity improves performance. We also show that alignment methods enable smaller, open-source models (Llama-3.1 8B and Mistral-7B) to approach the compromise generation quality of Claude 3 Opus, the proprietary model used to generate the training data.  To extend this work, we hope to explore socially intelligent tasks beyond place-based conflicts, including ethical or policy disagreements, and develop methods that further maximize empathy toward both viewpoints. This would help to broaden the framework's applicability to real-world mediation scenarios.


\clearpage

\bibliography{custom}

\clearpage
\appendix

\section{Limitations}

The COT+feedback approach sought to produce compromise text that is equally acceptable to both viewpoints. In the future, we would also like to include consideration of maximizing the empathy of each viewpoint to the compromise.

Our collection of compromises was constrained to keep the data collection required manageable. The compromises are generated based on two different human viewpoints for a similar type of place.  Although many of the comments about a place apply generally to a place type, \ie trails and fencing in a park, in a real scenario, the expressed viewpoints are about the same type of place, rather than exactly the same. We chose this approach because collecting compromise data for the same place would be a much larger undertaking. The prompts were based on feeling safe/unsafe or welcome/excluded. Generalization to other prompts was not done because this would have also increased the required size of the negative viewpoints. These generalizations are left for future work.
\par
While this place based focus on safety and inclusivity enables a controlled exploration of empathy-neutral compromises, it inherently limits the framework's immediate generalization to broader real-world scenarios, such as interpersonal conflicts, policy debates, or collaborative decision-making in non-spatial domains (\ie ethical dilemmas in team projects or ideological disagreements in community forums). The reliance on physical "places" as the locus of conflict may overlook nuances in abstract or multifaceted disputes, potentially reducing the framework's utility for diverse mediation contexts where viewpoints diverge on intangible issues.\par

To enhance generalizability, future work could expand the dataset to include contrasting viewpoints on non-place-based topics, such as political policies (\ie environmental regulations) or social norms (\ie workplace diversity initiatives), by scaling negative viewpoint collection through crowdsourcing platforms or synthetic augmentation techniques. This would test the framework's robustness across domains, potentially incorporating multi-turn dialogues to simulate ongoing negotiations.\par

\par

While our neutrality metric captures empathic balance, it does not account for societal biases (\ie race, gender, class) present in inputs or LLM generations. LLMs may overlook power asymmetries or ethical subtleties due to limited social reasoning. Moreover, the dataset’s narrow domain limits generalizability. Future work should explicitly evaluate and mitigate such biases.

The Human Study comparing different compromise generation texts presented only a sample of 4 of the generated texts per viewpoint pair, rather than all 16, because humans would have difficulty rating all items reliably.

\section{Ethical Considerations}
Our goal in this work was to generate compromises that are equally acceptable to the two parties with different viewpoints. Similar to the general dangers of some people trusting all results generated by an LLM, the generated compromises should be viewed as suggestions for discussion areas, rather than a compromise that each person should agree to.
\par
Our work is a first step . In the future, it should be used as an aid rather than being relied upon directly. However, a possible future misuse is in sensitive conflict scenarios, where the compromises are not used as suggestions but instead as direct solutions. This is a societal problem, similar to other issues that generative AI has created, such as the generation of fake videos of real people and the generation of mental health advice.

\section {Summary of Dataset Collection Method in \citet{chen2025empathy}}

\label{sec:dcm}
\vspace{-0.1em}
\begin{figure*}[!ht]    
    \centering
    \includegraphics[height=0.20\textheight, keepaspectratio]{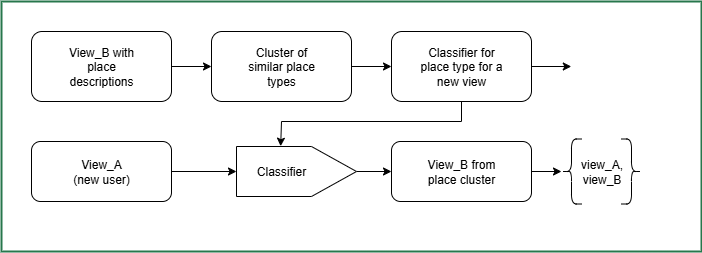}    
    \caption{Data collection methodology}
    \label{fig:datapointe}
\end{figure*}
\vspace{-0.01em}

We briefly summary the data collection method used by \citet{chen2025empathy}, which provided the contrasting viewpoints used in our work. As shown in Figure \ref{fig:datapointe}, in stage 1 (upper row), participants recruited from dscout\footnote{\url{https://dscout.com/}} provide views (view\_B) about places where they felt less safe/excluded. For each place, they explained their feelings and suggested modifications to improve safety or inclusivity. Place descriptions for less safe/excluded places were each grouped into 14 clusters using agglomerative clustering. A classifier was developed to match new views to these clusters.

In stage 2 (bottom row), new participants from Prolific \footnote{\url{https://www.prolific.com/}} wrote views (view\_A) about places where they felt safe/welcome. The classifier matched each view\_A to a view\_B from the place-type cluster corresponding to view\_A.

\section{Collection of Human Ratings for Evaluation of Prompting Strategies}
\label{app:humanEval}
The Prolific\footnote{\url{https://prolific.com}} platform was used to recruit participants located in the US. The participants were paid the equivalent of \$17/hr for 20 minutes of time, with the actual mean time of a batch of participants ranging from 14-20 minutes. All participants provided informed consent prior to participating in the study. The study was part of a larger study that was reviewed by a review board prior to launch of the study. A total of 50 participants were recruited for this part of the study.
  
The participants were given the following overall instructions:

\begin{quote}
Welcome to the study.
In this study, you will read accounts about similar places from different pairs of people, whom we will be labelling as Person A and Person B. The accounts are related to their perceptions of how welcoming or safe their neighborhoods are. A and B have not met and wrote their stories separately.

After reading each pair, you will take the perspective of Person A and rate the acceptability of several suggested modifications. How acceptable would Person A find these suggestions?
\end{quote}

Each participant was asked to rate compromises for five pairs of statements about suggested place modifications written by Person A and Person B. For each participant, the story A of a pair was first presented:
\begin{quote}
You are about to read Person A's story. Please pay special attention to this story. You will need to take this person's perspective when considering Person B's story and the suggestions afterwards.

You should have just read the story of Person A, the person whose perspective you will take.
\end{quote}

Whether story A is a positive view or a negative view was randomly assigned to participants, balancing for an equal number of positive and negative views.
Next, the participant read story B of a pair of statements:
\begin{quote}
Now you will read Person B's story. Person B has had a much different experience. (A and B have not met and wrote their stories separately).

Press the blue button when you are ready.
    
\end{quote}

After reading the stories by Person A and Person B, a participant is asked to rate four generated compromises plus Person B's modifications.

\begin{quote}
Now please pretend to be Person A (the first person whose story you read).

As Person A, you have just read Person B's story. You will see a list of suggested modifications that try to address both A and B. Carefully consider each of the following suggested modifications and rate how acceptable each suggestion is from your (Person A's) perspective. To help you remember, we have included both people's original suggestions below.

A's suggested modification: <view A>
B's suggested modification: <view B>

Pretend you are Person A (the first person whose account you read). 

How acceptable is each suggestion? Use the slider BELOW each suggestion to make your rating.
\end{quote}
The four suggestions plus view B were presented in random order, and a horizontal slider from 1 to 100 was shown below each text. 

Once all sliders were adjusted, the slider ratings were recorded and the participant could click a button to move to the next pair of statements.

Once the participant rated five pairs of statements, they were asked demographic questions about their age, gender, income level, education, ethnicity, and people in their household. We limited each participant to rating compromises for only five pairs of viewpoints because feedback from a pilot study indicated that participants felt that doing more would be tiring and less reliable.

\section{Single prompt}
\label{sec:sprompt}

\begin{tcolorbox}[title=Basic Single Prompt,width=\columnwidth, fontupper=\footnotesize,breakable,bottom=4mm]
You are an intelligent AI assistant! \par

I need you to generate a third person response strictly based on two contrasting views called positive story and negative story. \par

The positive and negative story should be equally empathetic towards the response. The response should be a specific suggestion. It should be a compromise between the positive and negative stories based on the context of both stories. \par

Please generate 4 responses with a fixed format. Try to be as specific and short instead of being comprehensive. \par

Please provide your response in the following format: \par
Positive view: \{\} \par
Negative view: \{\} \par
Response 1: [Insert response 1 here] \par
Response 2: [Insert response 2 here] \par
Response 3: [Insert response 3 here] \par
Response 4: [Insert response 4 here] \par
\end{tcolorbox}

\section{COT+LLM Prompt Template}
\begin{tcolorbox}[title=Prompt template,width=\columnwidth, fontupper=\footnotesize,breakable,bottom=4mm]

\textbf{Step 1:} Identify the suggestions in view\_A.

\vspace{2mm}

\textbf{Step 2:} Identify the suggestions in view\_B.

\vspace{2mm}

\textbf{Step 3:} Identify similarities in suggestion between view\_A and view\_B.

\vspace{2mm}

\textbf{Step 4:} Create N empathically neutral compromises using the similarities.

\vspace{2mm}

\textbf{Step 5:} Estimate empathy similarity scores between each compromise and view\_A/view\_B, respectively.

\vspace{2mm}

\textbf{Step 6:} Create N better response with higher empathy similarity score.

\end{tcolorbox}

\section{Iterative Prompt Engineering Procedure}

This section describes the iterative prompt engineering procedure used to generate collected compromises guided by a similarity model. The process consists of a fixed prompt component (computed once) and an iterative refinement loop that reuses these fixed outputs.

\subsection{Fixed Prompt Component (Computed Once)}

The following prompt steps are executed once at the beginning, and their outputs are cached and reused throughout all iterations:

\begin{tcolorbox}[title=Fixed Prompt - Computed Once,width=\columnwidth, fontupper=\footnotesize,breakable,bottom=4mm,colback=gray!10]

\textbf{Output 1:} Identify the suggestions in view\_A

\vspace{2mm}

\textbf{Output 2:} Identify the suggestions in view\_B

\vspace{2mm}

\textbf{Output 3:} Identify similarities in suggestion between view\_A and view\_B

\vspace{2mm}

\textit{Note: These outputs (suggestions\_A, suggestions\_B, similarities) are fixed and reused in all subsequent iterations.}

\end{tcolorbox}

\subsection{Step 1: Initial Compromise Generation}

Using the fixed outputs from above, generate initial compromises:

\begin{tcolorbox}[title=Generation Prompt - Step 1,width=\columnwidth, fontupper=\footnotesize,breakable,bottom=4mm]

\textbf{Input:} suggestions\_A, suggestions\_B, similarities (from fixed prompt)

\vspace{2mm}

\textbf{Prompt:} Create N empathically neutral compromises using the similarities

\vspace{2mm}

\textbf{Output:} Compromise\_1, Compromise\_2, ..., Compromise\_N

\end{tcolorbox}

\vspace{2mm}

The generated compromises are then evaluated by the Similarity model using Claude API, which produces empathy similarity scores (\texttt{score\_A}, \texttt{score\_B}) for each compromise with respect to view\_A and view\_B.

\subsection{Step 2: Iterative Refinement}

Using the same fixed outputs, iteratively refine the compromises:

\begin{tcolorbox}[title=Refinement Prompt - Step 2 (Iterative),width=\columnwidth, fontupper=\footnotesize,breakable,bottom=4mm]

\textbf{Input:} 
\begin{itemize}[nosep,leftmargin=*]
    \item suggestions\_A, suggestions\_B, similarities (from fixed prompt)
    \item compromise (from previous iteration)
    \item score\_A, score\_B (from similarity model)
\end{itemize}

\vspace{2mm}

\textbf{Prompt:} Create N better responses with higher empathy similarity score

\vspace{2mm}

\textbf{Output:} Refined\_Compromise\_1, ..., Refined\_Compromise\_N

\end{tcolorbox}

\vspace{3mm}

\noindent\textbf{Iterative Process:} Each refined compromise is evaluated by the Similarity model to obtain updated scores. The compromise with improved scores can be further refined by repeating Step 2, while always reusing the fixed prompt outputs (suggestions\_A, suggestions\_B, similarities) computed in the initial step. This ensures consistency across iterations while allowing the generation component to focus on improving empathy similarity scores.

\section{Qualitative analysis of candidate compromise generation across different prompting strategies}

\label{sec:failure}

We provide a detailed qualitative example of target compromise generation across different prompting strategies. The first example provides a comprehensive evaluation, while the second example features a brief assessment.
 \label{sec:aprompt}

\begin{tcolorbox}[title=Example 1,width=\columnwidth]

\textbf{Positive view:}
I am writing about this place: A nearby park. I feel safe here because I feel safe here when others are around. There's a good sense of community. Some ways this place could be modified to be safer are: There's no fences, gates, no visitor check, and it's extremely open. This is good and bad.

\textbf{Negative view:}
I am writing about this place: Our local park. I feel safety could be improved here. It's a wonderful park, but people do not obey leash laws. I often see people letting their dogs off leash, making other dog walkers and people with children uncomfortable. Some ways this place could be modified to be safer are I would like to see stricter leash laws. Fines for rule breakers.
\end{tcolorbox}

\subsection*{Results}
\subsection*{Single Prompt}

\begin{itemize}
\item Response 1: \textcolor{darkred}{The park management could consider implementing designated off-leash areas within the park, allowing dogs to roam freely in specific zones while maintaining strict leash laws in other areas to ensure the comfort and safety of all visitors.}

\item Response 2: \textcolor{darkred}{ Increasing the presence of park rangers or volunteers to monitor and enforce leash laws could help maintain a safer environment for everyone, while also providing an opportunity to educate dog owners about the importance of responsible pet ownership.}

\item Response 3: \textcolor{darkred}{The park could install clear signage outlining the leash laws and the consequences for violating them, such as fines or temporary bans from the park, to encourage compliance and create a safer atmosphere for all visitors.}

\item Response 4: \textcolor{darkred}{ Organizing community events that promote responsible dog ownership and provide opportunities for dogs and their owners to socialize in a controlled environment could help foster a sense of community and encourage adherence to park rules and regulations.}

\end{itemize}
\subsubsection*{Analysis}
\begin{itemize}

\item Human observation: The basic single prompt strategy shows clear limitations in balanced perspective handling. Claude's responses predominantly focus on addressing the negative story's concerns about dogs and leash laws, while largely overlooking the positive story's suggestions about gene ral safety and community aspects. This demonstrates a bias in compromise generation toward more concrete, specific issues (dog-related) over broader community safety concerns.

\end{itemize}
\subsection*{Chain of Thought(CoT)}

\subsubsection*{Results}
\begin{itemize}
\item Response 1:  \textcolor{lightred}{Install fences and gates around designated off-leash areas for dogs, ensuring they are separate from other park visitors.}
\item Response 2: \textcolor{lightred}{Implement a visitor check system that includes registration for dog owners, ensuring they are aware of and agree to follow leash laws.}
\item Response 3: \textcolor{lightred}{Increase park staff presence to enforce leash laws and issue warnings or fines to rule breakers, promoting a safer environment for all visitors.}
\item Response 4: \textcolor{lightred}{Add clear signage throughout the park to remind visitors of leash laws and the importance of maintaining a safe and comfortable environment for everyone.}
\end{itemize}
\subsubsection*{Analysis}
\begin{itemize}

\item Human Observation: The Chain of Thought approach shows improvement in addressing multiple aspects of safety, but still demonstrates some imbalance. While responses successfully incorporate the positive story's suggestions about physical safety measures, the solutions tend to be more rigid and enforcement-focused.
\end{itemize}
\subsection*{CoT+Feedback}

\subsubsection*{Results}
\begin{itemize}
\item Response 1:\textcolor{darkgreen}{ Create designated off-leash areas within the park that are securely fenced and have separate entrances. These areas should be located away from the main paths and gathering spaces to minimize potential conflicts between dogs and other park visitors. Clearly mark the off-leash areas and provide amenities such as water fountains and waste disposal stations.}

\item Response 2: \textcolor{darkgreen}{ Implement a graduated approach to park safety, starting with clear signage outlining park rules, leash laws, and consequences for violations. Provide educational materials on responsible pet ownership and the importance of maintaining a safe environment for all visitors. As a second step, strategically install fences and gates in high-traffic areas to improve safety while preserving the park's openness. Regularly assess the effectiveness of these measures and adjust as needed based on community feedback.}

\item Response 3: \textcolor{darkgreen}{Implement a hybrid approach to park safety that combines educational initiatives with targeted infrastructure improvements. Begin by launching a comprehensive awareness campaign that encourages responsible pet ownership, highlights the importance of leash laws, and promotes a culture of mutual respect among all park users. Simultaneously, identify specific areas within the park where safety concerns are most prevalent and install strategic fencing and gates to mitigate risks without compromising the overall openness of the space. Regularly assess the effectiveness of these measures through community surveys and data analysis, making adjustments as needed to strike a balance between safety and accessibility.}

\item Response 4: \textcolor{darkgreen}{Implement a "Balanced Park Safety" initiative that addresses the concerns of both park user groups. The initiative will focus on creating a safe and welcoming environment for all visitors, including those with children and dog owners. To achieve this, the park will introduce designated off-leash areas for dogs, ensuring that they have a secure space to play and socialize without compromising the safety and comfort of other park users. Simultaneously, the park will enforce leash laws in all other areas, with clear signage and regular patrols by park staff to encourage compliance. The initiative will also include educational programs for dog owners, promoting responsible pet ownership and emphasizing the importance of respecting park rules and other users\' needs.}

\end{itemize}
\subsubsection*{Analysis}
\begin{itemize}
\item Human Observation: The Chain of Thought with Feedback approach demonstrates the most comprehensive and balanced response generation. All five responses effectively address the negative story's dog-related safety concerns while incorporating the positive story's suggestions about physical safety measures. Notably, this approach generates creative solutions beyond the original suggestions, such as community feedback mechanisms and educational initiatives. The responses show deeper integration of both perspectives and introduce novel elements that enhance both safety and community aspects, demonstrating superior compromise generation capabilities.
\end{itemize}
\newpage

\begin{tcolorbox}[title=Example 2]

\subsubsection*{Positive view}
 I am writing about this place: This is our local church. I feel welcomed by others for who I am in this location because This church has become our home church after our previous church split up. They readily took our family in and made us feel loved and welcome..Some ways this place could be modified to be more welcoming are Honestly, I would not change anything about the practices, but I might meet with them to find out why they feel excluded. I would give them the gospel.

\subsubsection*{Negative view}
I am writing about this place: Church,Downtown, Vegas. I feel excluded by others for who I am in this location because Unfortunately, organized religion and the LGBT+ community for the most part live in a space like oil and water. While there are some exceptions, many religious institutions utilize their beliefs as weapons as opposed to a welcome. I was raised as a catholic, but as I became an adult I quickly realized that who I am does not align with the beliefs and practices within organized religion. This space never feels welcoming or inclusive. It's a space I focus on avoiding with all of my energy. The sad part is the roots of religion are based off inclusion and making those who are looked upon by society as “less than” feel lifted up and included, however, that is not what is put into practice today..Some ways this place could be modified to be less excluding and more welcoming are If the church and  organized religion went back to core principals of love, praise and inclusion, I truly feel that a space such as this beautiful church would be one that I would walk to weekly to share in a joyous occasion. It's unfortunate because this is exactly a half block from where I live. I hope one day that there is the realization that have veered off the right path. It takes one sector of organized religion to start an inclusion practice and accept no exceptions.
\end{tcolorbox}
\subsection*{Single Prompt}

\subsubsection*{Results}
\begin{itemize}
\item Response 1: \textcolor{darkred}{Design an 'Inclusive Community Framework' with designated quiet zones and gradual integration options, using a holistic screening approach that considers present circumstances over past history, accommodating both those who prefer minimal interaction and those seeking community acceptance.} [\textcolor{navyblue}{Framework too complex and potentially overwhelming}]

\item Response 2:  \textcolor{darkred}{Institute a 'Neighborhood Integration Choice' system where residents can customize their community involvement level, with fair entry requirements focusing on future contribution rather than past circumstances.} [\textcolor{navyblue}{Integration choice system may reinforce divisions}]

\item Response 3: \textcolor{darkred}{Create a 'Neighborhood Harmony Program' featuring both private sanctuaries and optional community spaces, with flexible residency requirements that focus on present community contribution and mutual respect, accommodating both those seeking solitude and those working on personal growth.} [\textcolor{navyblue}{Harmony program too bureaucratic and impersonal}]

\item Response 4: \textcolor{darkred}{Create community integration programs that provide both structured support for newcomers (including those in recovery) and optional private participation methods, respecting individual comfort levels with social interaction.} [\textcolor{navyblue}{ Community integration focus may miss core issues}]

\end{itemize}

\subsection*{CoT}

\subsubsection*{Results}

\begin{itemize}
\item Response 1: \textcolor{lightred}{Institute monthly community meetings where both current church members and those who feel excluded can work together to define specific, actionable steps toward mutual understanding and respect.} [\textcolor{navyblue}{ Community meetings might lack sufficient structure}]

\item Response 2: \textcolor{lightred}{Develop a "Welcome All" program that combines gospel sharing with inclusive practices, allowing different groups to participate in church community service projects together.} [\textcolor{navyblue}{"Welcome All" program could feel pressuring to both sides}]

\item Response 3: \textcolor{lightred}{Establish regular open dialogue sessions where church members and those feeling excluded can share their perspectives in a structured, respectful environment focused on finding common ground.} [\textcolor{navyblue}{Open dialogue without clear goals may not progress}]

\item Response 4: \textcolor{lightred}{Create a dedicated outreach committee that includes both traditional church members and LGBTQ+ individuals to discuss ways to bridge understanding while maintaining church values.} [\textcolor{navyblue}{Outreach committee needs more defined purpose}]

\end{itemize}

\subsection*{CoT+LLM score}

\subsubsection*{Results}

\begin{itemize}
\item Response 1: \textcolor{amber}{The church could partner with local LGBTQ+ organizations to develop inclusive worship services that maintain religious traditions while incorporating elements that acknowledge and celebrate diverse identities, perhaps on alternate weekends.} [\textcolor{navyblue}{Partnership for inclusive services is good but may push too far too fast}]

\item Response 2: \textcolor{amber}{The church leadership could implement monthly "Open Door Sundays" where they explicitly welcome and engage with diverse community members through structured interfaith discussions, followed by shared meals to build personal connections.} [\textcolor{navyblue}{Open Door Sundays provides structure but might feel forced}]

\item Response 3: \textcolor{amber}{Create specific inclusion guidelines that combine traditional church values with explicit welcoming practices, developed collaboratively by church leadership and community members.} [\textcolor{navyblue}{ Inclusion guidelines are helpful but may be difficult to implement fairly}]

\item Response 4: \textcolor{amber}{Create a dedicated outreach committee that includes both long-term members and those who have felt excluded, meeting monthly to discuss inclusion challenges and solutions.} [\textcolor{navyblue}{Outreach committee has potential but needs careful balance}]

\end{itemize}

\subsection*{CoT+Feedback}
\subsubsection*{Results}

\begin{itemize}
\item Response 1: \textcolor{darkgreen}{The church could create a dedicated safe space program where trained mediators host bi-weekly sessions allowing both traditional church members and LGBTQ+ individuals to share their perspectives and find common ground in their faith journey.} [\textcolor{navyblue}{Respecting both traditional religious practice and LGBTQ+ perspectives}]

\item Response 2: \textcolor{darkgreen}{The church could establish an LGBTQ+ liaison committee composed of both congregation members and LGBTQ+ community representatives to facilitate meaningful dialogue and identify specific inclusion opportunities while respecting core religious values.} [\textcolor{navyblue}{Focuses on practical solutions}]

\item Response 3: \textcolor{darkgreen}{Establish regular open dialogue sessions where church members and those feeling excluded can share their experiences and perspectives in a moderated, respectful environment.} [\textcolor{navyblue}{Regular open dialogue sessions work well as they allow both sides to share experiences}]

\item Response 4: \textcolor{darkgreen}{Develop a structured welcome program where new or returning members are paired with existing church members for support and integration, regardless of their background.} [\textcolor{navyblue}{Being inclusive while respecting church structure}]

\end{itemize}

\section{Qualitative analysis of compromise generation}
\label{sec:quality}
\begin{tcolorbox}[title=Example 1 (safe/less safe pair)]

\subsubsection*{Positive view}
 I am writing about this place: The river next to a trail that I walk on. I feel safe here because I feel safe because it brings me joy and helps me to feel good. I feel comforted and comfortable and therefore feel safe.. Some ways this place could be modified to be safer are :It may feel safer knowing that people are around if your personal safety is a concern. There is only one path and no ability to get lost. Everyone is friendly.

\subsubsection*{Negative view}
I am writing about this place: The river walk I feel safety could be improved here There have been several attacks along this river walk. It's not a place that people want to enjoy after dark. Part of the problem is the amount of drunk and high people that frequent the path after dark..Some ways this place could be modified to be safer are While there are lights and it is somewhat illuminated, it would be helpful to have regular security guards along the trail at night. Security cameras would also be very helpful along this walkway.
\end{tcolorbox}

\subsubsection*{Generation}

\textbf{Base model (Llama):} The riverside trail is fundamentally a place of community and connection. By establishing regular morning walking groups between 7-9 AM, we can enhance the existing friendly atmosphere. The natural gathering of people who enjoy the path creates an organic safety network. The single clear path already provides good navigation, and scheduled group activities would further strengthen the welcoming environment that makes this place special. [\textcolor{navyblue}{ Leans heavily towards Positive view's  perspective.}]

\textbf{Base model (Llama)+FT:} The river walk requires enhanced security measures to ensure safety at all hours. Implementing a dedicated patrol system with trained local volunteers would provide necessary oversight, especially during evening hours. Installing emergency communication points and improving lighting would address the current safety concerns. These measures would help transform the area from a potentially dangerous space into a secure community asset. [\textcolor{navyblue}{ Focuses heavily on Negative view's security concerns.}]

\textbf{Aligned Llama:} Create a "Community Safety Network" that combines the trail's natural friendliness with practical security measures. This would include trained volunteers from the local community who patrol during both daylight and evening hours, wearing recognizable but approachable uniforms. Establish welcoming rest areas to serve as both social gathering points and safety zones, encouraging community interaction. [\textcolor{navyblue}{ Maintaining balanced consideration of both perspectives.}]

\begin{tcolorbox}[title=Example 2 (welcome/excluded pair)]

\subsubsection*{Positive view}
 I am writing about this place:  North Park, in my town.. I feel welcomed by others for who I am in this location because It's a hidden park that is rather unknown, it's always quiet, has a bench by the creek and trails to walk. I feel welcomed here because it allows me to take a break from my day and decompress.Some ways this place could be modified to be more welcoming are Someone who is not open to mindfulness, or who is loud and doesn't like being alone, may dislike this place. I would imagine they may feel too alone with their thoughts, and not able to appreciate what is around them due to what they may already be used to elsewhere, perhaps they thrive in a city environment. I would make this space more welcoming for them, simply by accompanying them on their visit. Providing them comfort while they experience what I do while I alone.

\subsubsection*{Negative view}
I am writing about this place: The park that is in our neighborhood. I feel excluded by others for who I am in this location because I went a few times with my son and the people that have been there each time that we have gone pretty much claimed it for themselves. I actually know them from the school that I teach at and they don't like me because I am a Christian. At one point they said "um we are playing here" , And you could tell by tone of voice and facial expression they didn't want us there. You can find imagine that would make someone not even want to stay.Some ways this place could be modified to be less excluding and more welcoming are I would maybe put a sign up saying all people are welcome in our neighborhood or community. Not that that would change much but it would at least make me feel better
\end{tcolorbox}

\subsubsection*{Generation}

\textbf{Base model (Mistral-7b):} The park should establish dedicated quiet zones with meditation benches and nature trails, creating peaceful spaces for individual reflection and mindfulness. Simple signs can indicate these areas as spaces for contemplation, helping preserve the serene atmosphere while gently encouraging respectful behavior from all visitors [\textcolor{navyblue}{ minimally addressing negative view's concerns about exclusion.}] \\

\textbf{Base model (Mistral-7b)+FT:} The park should implement a welcoming committee program where community volunteers greet visitors and organize inclusive group activities. Installing welcome signs in multiple languages and creating designated community gathering spaces would help ensure everyone feels accepted, regardless of their background or beliefs [\textcolor{navyblue}{ No consideration for those who find peace in solitude}] \\

\textbf{Aligned Mistral-7b:} Clear signage and designated spaces would establish areas for both group activities and individual reflection, supported by a comprehensive community engagement plan addressing personal and social needs. A trained community mediator would help resolve any conflicts or exclusion, while regular forums bring diverse groups together to shape park policies that give equal weight to all experiences and concerns. [\textcolor{navyblue}{ Maintaining balanced consideration of both perspectives.}] \\

\subsection*{Conclusion}
The Chain of Thought with Feedback approach generated the most balanced and nuanced responses, effectively integrating perspectives from both the positive and negative stories. It introduced creative solutions that enhanced both safety and community aspects. While this strategy has limitations such as being resource-intensive and potentially introducing bias, it demonstrated superior compromise generation capabilities compared to the basic single prompt and Chain of Thought approaches.

\section{Significance of Demographics}
\label{sec:demo}
To evaluate the potential impact of demographic context on compromise generation, we explored whether it benefits from demographic information provided as input. Two base models were independently fine-tuned and aligned: one with demographic data included, and one without. Each model was trained on 300 pairs of contrasting views from the training set and evaluated on 50 pairs from the test set. Incorporating demographic features (Table~\ref{tab:ab1})  yielded insignificant improvements, likely due to the task's emphasis on perceptual empathy over identity based cues. This  result aligns with prior empathy prediction work~\cite{
hasan2024llm,guda2021empathbert} reinforcing that viewpoint-specific metrics suffice for LLM compromise generation.

\vspace{-0.4em}
\begin{table}[h]
\centering
\vspace{5pt} 
\small
\begin{tabular}{l cc}
\hline
Experiment & \textbf{Llama}($\uparrow$) & \textbf{Mistral-7b}($\uparrow$) \\
\hline
\textbf{w demog.} & 0.249 & 0.238 \\
\textbf{w/o demog.} & 0.251 & 0.237 \\
\hline
\end{tabular}%
\caption{Quantitative analysis (ROUGE-L) on the effect of demographics as input. \textbf{w demog.} includes both the views along with the demographics of the person, While \textbf{w/o demog.} does not include demographics.Llama = Llama+FT+NCE, Mistral-7b = Mistral-7b+FT+NCE.}
\label{tab:ab1}
\end{table}
\vspace{-0.5em}

\begin{table}[h]
\centering
\small
\begin{tabular}{l r}
\hline
\textbf{Model} & \textbf{LL ($\uparrow$)} \\
\hline
LLAMA+DPO & -1.9145 \\
Mistral-7b+DPO & -1.8976 \\
Mistral-7b+NCE & -1.7230 \\
Mistral-7b+Task-loss based & -1.7015 \\
LLAMA+NCE & -1.6234 \\
LLAMA+Task-loss based  & -1.6345 \\
\hline
\end{tabular}%
\caption{Catastrophic forgetting test. LL: log-likelihood}
\label{tab:ab2}
\end{table}

\section{Catastrophic Forgetting}
\label{subsec:cata}
To evaluate the impact of our alignment approach on pretrained knowledge retention, we assess how well the model preserves its capabilities. We use the WikiText test set (from WikiText-2-v1) \footnote{\url{https://huggingface.co/datasets/Salesforce/wikitext/viewer/wikitext-2-v1/test}} \cite{merity2018scalable}, a widely used benchmark for measuring catastrophic forgetting \cite{fawi2024curlora}. We compute the average log-likelihood\footnote{Log-likelihood is computed by summing the log probabilities of all tokens in a datapoint.} across the dataset. A decrease in log-likelihood indicates that the model is forgetting previously learned knowledge. As shown in Table~\ref{tab:ab2}, our method better preserves pretrained knowledge compared to prior approaches, such as Direct Preference Optimization (DPO) \cite{rafailov2024direct}. This retention is critical for generating compromises that rely on the model's foundational understanding.

\section {Task-loss based alignment }
\label{sec:task_loss}

The NCE objective (~\ref{subsec:nces}) teaches the model to distinguish good compromises from poor ones 
similar to to how exposure to norm-conforming and norm-violating behavior builds evaluative standards~\cite{bicchieri2005grammar}.
 However, recognizing a good compromise is not the same as producing one. 
  In human mediation, the ability to generate fair proposals requires more than ranking. it requires an internalized sense of what balanced proposals substantively contain. Similarly, discrimination alone can drift in a model setting \ie a model may learn to assign high likelihood to outputs that are stylistically similar to good compromises without preserving their semantic content.
 
To capture this, we introduce a complementary task-loss objective that enforces not only preference ordering but also semantic fidelity to the validated compromise.
\vspace{-0.2em}
\begin{equation}
\label{eq:tbl}
\begin{split}
\mathcal{L}_{\text{tbl}} = \max(0, s_{hypo} - s_{target} + \\
|target_{rouge} - hypo_{rouge}| \cdot w_{\text{margin}})
\end{split}
\end{equation}

The quantities $s_{target}$ and $s_{hypo}$ denote the model’s scores (log-likelihoods) assigned to the target compromise and the generated hypothesis, respectively. 
The required preference gap is determined by the ROUGE difference. The loss encourages the target and the reference compromise and remain semantically close to it.  Because both scores are produced by the same model parameters, increasing the likelihood of the target also makes the model more likely to generate target-like hypotheses over time. 

In this formulation, $target_{rouge}$ denotes the ROUGE score of the candidate compromise with respect to itself (used as reference), while $hypo_{rouge}$ is the ROUGE score of the fine-tuned model’s generated hypothesis with respect to the corresponding candidate compromise. The term $w_{\text{margin}}$ (weight margin) modulates the influence of ROUGE difference during training. Note that while ROUGE is non-differentiable, our objective avoids direct backpropagation through the metric. The term $(target_{rouge} - hypo_{rouge})$ is computed during the forward pass and treated as a static scalar margin detached from the computation graph. This effectively scales the gradient applied to the log-probabilities, penalizing hypotheses proportional to their semantic deviation from the target without requiring differentiable approximations.

\end{document}